\documentclass{llncs}
\usepackage{graphicx} 
\usepackage{booktabs}
\usepackage{algorithm2e}

\begin{document}

\title{Deep Super Learner: \\ A Deep Ensemble for Classification Problems
}

%
%
\author{Steven Young, Tamer Abdou, and Ayse Bener }
%
%
%
\institute{Data Science Laboratory, Ryerson University, Toronto ON M5B 2K3, Canada,\\
\email{\{steven.young, tamer.abdou, ayse.bener\}@ryerson.ca}}

\maketitle              

\begin{abstract} Deep learning has become very popular for tasks such as predictive modeling and pattern recognition in handling big data.  Deep learning is a powerful machine learning method that extracts lower level features and feeds them forward for the next layer to identify higher level features that improve performance.  However, deep neural networks have drawbacks, which include many hyper-parameters and infinite architectures, opaqueness into results, and relatively slower convergence on smaller datasets.  While traditional machine learning algorithms can address these drawbacks, they are not typically capable of the performance levels achieved by deep neural networks.  To improve performance, ensemble methods are used to combine multiple base learners.  Super learning is an ensemble that finds the optimal combination of diverse learning algorithms.  This paper proposes deep super learning as an approach which achieves log loss and accuracy results competitive to deep neural networks while employing traditional machine learning algorithms in a hierarchical structure.  The deep super learner is flexible, adaptable, and easy to train with good performance across different tasks using identical hyper-parameter values.  Using traditional machine learning requires fewer hyper-parameters, allows transparency into results, and has relatively fast convergence on smaller datasets.  Experimental results show that the deep super learner has superior performance compared to the individual base learners, single-layer ensembles, and in some cases deep neural networks.  Performance of the deep super learner may further be improved with task-specific tuning.

\end{abstract}

\footnotetext[1]{This paper was written as part of the Certificate in Data Analytics, Big Data, and Predictive Analytics at Ryerson University}

\section{Introduction}
Deep learning is a machine learning method that uses layers of processing units where the output of a layer cascades to be the input of the next layer and can be applied to either supervised or unsupervised learning problems~\cite{Bengio2013RepresentationPerspectives}~\cite{Langkvist14areview}. Deep neural networks (DNN) is an architecture of deep learning that typically has many connected units arranged in layers of varying sizes with information being fed forward through the network. DNN have been successfully applied to fields such as computer vision and natural language processing, having achieved accuracy rates similar or superior to humans in classification~\cite{Schmidhuber:2012:MDN:2354409.2354694}. For example, Ciresan et al. using DNN achieved an error rate half the rate of humans in recognizing traffic signs.  The multiple layers of a DNN allow for varying levels of abstraction and the cascade between the layers enables the extraction of features from lower to higher level layers to improve performance~\cite{Bengio:2009:LDA:1658423.1658424}. However, DNN also have drawbacks, listed below:
\begin{itemize}
\item DNN have many hyper-parameters, which are parameters where their values are set prior to training as opposed to parameter values that are set via training, that interact with each other in their relation to performance.  Numerous hyper-parameters, together with infinite architectures, makes tuning of hyper-parameter and architecture difficult~\cite{Zhou2017}.
\item With a large number of processing units, tracing through a DNN to understand the reasoning for classifications is difficult, leading to DNN being treated as black boxes~\cite{10.1162NECO00409}.
\item DNN typically require very large amounts of data to train and do not converge as fast, with respect to sample size, as traditional machine learning algorithms~\cite{Farrelly2017}. 
\end{itemize}
Traditional machine learning algorithms, on the other hand, are relatively simple to tune and their output may provide interpretable results leading to a deeper understanding of the problem, though they tend to underperform DNN in terms of accuracy.

The remainder of this paper is organized as follows: section 1 introduces the motivation and background for this paper, section 2 presents the overall procedure of the DSL approach, section 3 describes the methodology of the experiment, section 4 presents the results of a comparison of the performance of the DSL to the individual base learners and a selection of ensembles and DNN on various problems, and section 5 concludes and describes future work.


\subsection{Motivation}

Given the drawbacks of DNN and the poor performance of traditional machine learning algorithms in some domains and/ or prediction tasks, this paper investigates whether traditional machine learning algorithms can be used to address the drawbacks of DNN and achieve levels of performance comparable to DNN.  A new ensemble method, named here as Deep Super Learner (DSL), seeks to have simplicity in setup, interpretability of results, fast convergence on small and large datasets with the power of deep learning.  

\subsection{Ensemble Methods}

Ensemble methods are techniques that train multiple learning algorithms, which in combination yields significantly higher accuracy results than a single learner~\cite{Seni2010}.  Common methods include boosting, bagging, stacking, and a combination of base learners.  Each of these methods are tested for performance in this paper.  Boosting takes a model trained on data and incrementally constructs new models that focus on the errors in classifying made by the previous model.  An example is XGBoost, which is an efficient implementation of gradient boosting decision trees~\cite{Chen:2016:XST:2939672.2939785}.  Bagging involves training models on random subsamples and then each model votes with equal weight on the classification.  Random forest uses a bagging approach to enable the selection of a random set of features at each internal node~\cite{Xie:2009:CBB:3000364.3000367}.  Stacking takes the output of a set of models and feeds them into another algorithm that combines them to make the final predictions.  Any arbitrary set of base learners and combiner algorithm can be used.  Combination takes the predictions of the models and combines them with a simple or weighted average.  Super learner is a combination method that finds optimal weights to use when calculating the final prediction~\cite{VanderLaan2007}.

Super learning is an ensemble method proposed by Van der Laan et al. that optimizes the weights of the base component learners by minimizing a loss function given cross-validated output of the learners. Super learning finds the optimal set of weights for the learners and guarantees that performance will be at least as good as the best base learner~\cite{VanderLaan2007}.  The proposed algorithm, DSL, is an extension of the super learner ensemble.

When constructing an ensemble, having diversity among the component learners is essential for performance and a strong generalization ability~\cite{Zhou:2012:EMF:2381019}.  The super learner adapts to various problems given a set of diverse base learners since the weights of the components are optimized for the problem as different learners perform differently on different problems. There is also flexibility in the set of base learners to use depending on requirements or constraints as dictated by the problem or computational resources.

\subsection{Related Work}
\vspace{-1mm}
Very little research has been conducted on using traditional machine learning in a deep learning architecture.  Zhou and Feng describe a tree-based deep learning architecture~\cite{Zhou2017}.  However, the use of only decision tree based algorithms and the use of a simple average to combine the results of the base learners may limit the ultimate performance of this approach and its adaptability to diverse set of problems.  Farrelly tested an architecture using traditional learners arranged in three hidden layers~\cite{Farrelly2017}.  It is unclear if the implementation allowed iteration to continue the deep learning.  Accuracy results from this architecture did not outperform the super learner.

\section{The Proposed Deep Super Learner Approach}
\begin{figure}[!htp]
\centering 
\includegraphics[width=13cm]{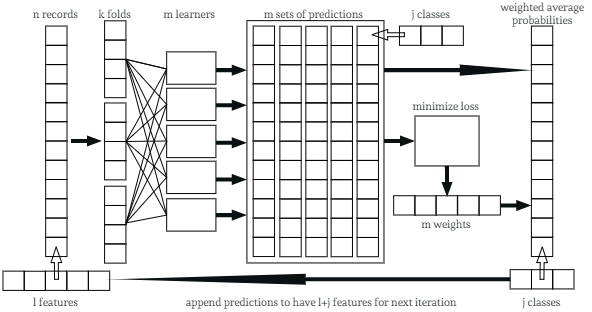}
\caption{Overall procedure of DSL with $j$ classes, $k$ folds, $l$ features, $m$ learners, $n$ records} 
\label{fig:1} 
\vspace{-5mm}
\end{figure}

\vspace{-1mm}
Deep learning consists of a layer by layer processing of features with a cascading hierarchy structure where the information processed by a layer is fed to the next layer for further processing.  The deep super learner essentially uses a super learning ensemble for each layer.  The overall training process and hyper-parameter values used are described below (see Figure~\ref{fig:1} and Algorithm~\ref{alg:1}).  Let there be $j$ classes, $k$ folds, $l$ features, $m$ base learners, and $n$ records in the training set.

\begin{enumerate}
\item Cross-validation is used to generate out-of-sample predictions for the entire training set.  Split the training set into $k$ equal size, $n/k$, mutually exclusive groups to be used as validation sets.  For each of the $k$ validation sets form $k$ folds where all the $n-n/k$ training records not in the validation set are used for training for that validation set.  The number of folds can impact the degree of under and over-fitting of the algorithm as well as runtime since with a higher number of folds each fold contains a larger portion of the training data to train on leading to a better fit, all else being equal.  However, with more folds there is a greater overlap in the training data between the folds leading to potential over-fitting and more runtime.  Three folds are used in this paper as experimentation showed three folds to be a good balance of fit and runtime.
\item Build and train each of the $m$ base learning algorithms on each fold.  Each model outputs class probabilities for each fold.  There are $k*m$ trained models.  Bring the predictions on each validation set together for each learner to obtain predictions for the entire training set.  This paper uses five base learners:  logistic regression, k-nearest neighbors, random forest, extremely randomized trees, and XGBoost.  Logistic regression, k-nearest neighbors, and random forest are used as they represent three different classification models with different philosophies and performance on different datasets~\cite{Lessmann2008BenchmarkingFindings}.  Extremely randomized trees and XGBoost are used for additional diversity within tree-based approaches.  The hyper-parameters of these learners are described below.
\item Optimize a loss function to find the linear combination of predictions from each of the $m$ learners to minimize the objective against the true values for the training data.  Save the value of the loss function and the optimized weights.  Log loss, which is a convex function and therefore convex optimization is sufficient, is used here.
\item With the optimized weights, calculate the weighted average of the predictions across learners to obtain overall predictions for each record.
\item (Optional) Re-train each of the models on the entire training set to get $m$ trained models.  Super learning as described by Van der Laan et al. requires this step~\cite{VanderLaan2007}.  However, with a sufficient number of cross-validation folds, as described above, the trained models will be trained on a sufficient portion of the training data where additional training data does not improve goodness of fit. Re-training may also be computationally expensive.  This step is recommended when the number of folds is low or when making predictions is more computationally expensive than training, such as for k-nearest neighbors.  This step was not performed for this paper as experimentation showed no significant difference in performance on the tested datasets.
\item Append the overall predictions to the original training data.  For example, if there are $j$ classes, step 4 produces a $j$-dimensional vector containing class probabilities for each record.  These vectors are concatenated as additional features to the original l-dimensional feature vectors for the records, resulting in a total of $l+j$ features.
\item Feed the training data augmented with the predictions through the steps above.  Repeat this process until the optimized loss value no longer decreases with each iteration.  Save the number of iterations after which the training process ends.
\end{enumerate}

\begin{algorithm}[!h]\label{alg:1}
\SetAlgoLined
 \For{iteration in 1 to max iterations}{
  Split data into k folds each with train and validate sets\;
  \For{each fold in k folds}{
  		\For{each learner in ensemble}{
  			Train learner on train set in fold\;
			Get class probabilities from learner on validate set in fold\;
			Build predictions matrix of class probabilities\;
  			}
  		}
Get weights to minimize loss function with predictions and true labels\;
Get average probabilities across learners by multiplying predictions with weights\;
Get loss value of loss function with average probabilities and true labels\;
\eIf{loss value is less than loss value from previous iteration}{
	Append average probabilities to data\;
   }{
   	Save iteration\;
	Break \textbf{for}\;
  }
 }
 \caption{A Pseudo-code of the Proposed Approach, DSL}
\end{algorithm}

To make predictions on unseen test data, pass the data in its entirety through a similar process using each of the models trained and weights optimized at each iteration.  If the models are trained on the entire training set, use these $m$ models for each iteration.  If the models are trained on the $k$ folds, use each model trained on each fold to make predictions on all the unseen data and average across the $k$ models to get predictions for each of the $m$ learners.  Using the optimum weights for the $m$ learners found during training for the iteration, calculate the overall weighted average predictions for the iteration.  Append the predictions to the original test data as additional features.  Repeat the process for the same number of iterations used in training.

\section{Methodology}

The hyper-parameters and architectures for the DSL, base learners, benchmark ensembles, and benchmark DNN described below are kept constant between datasets.  When necessary, adjustments are made for the different dimensionality of the datasets.
\subsection{Base Learners and Benchmark Ensembles}
The same five base learners used in DSL are also tested individually and in the benchmark ensembles using identical hyper-parameter values.  If a hyper-parameter of a learner is not listed below, in Table~\ref{tab:1}, default values of the implementation of the algorithm are used.


\begin{table}[!ht]
\centering
\caption{Summary of hyper-parameters for base learning algorithms}
\label{tab:1}
\begin{tabular}{@{}p{5cm}l@{}}
\toprule
Base learner                                   & Hyper-parameters                                                                                                                         \\ \midrule
Logistic regression                            &  N/A                                                                                                                                      \\
k-Nearest neighbors                            & Neighbors: 11                                                                                                                            \\
Random forest                                  & Trees: 200; Depth: unlimited                                                                                                             \\
\multicolumn{1}{l}{Extremely randomized trees} & \begin{tabular}[c]{@{}l@{}}Trees: 200; Depth: unlimited; \\ Max features when splitting: 1\end{tabular}                                \\
XGBoost                                        & \begin{tabular}[c]{@{}l@{}}Trees: 200; Max depth: 3;\\  Row subsampling: 0; \\ Column subsampling: 0;  Learning rate: 1\end{tabular} \\ \bottomrule
\end{tabular}
\vspace{-5mm}
\end{table}

Since random forest, extremely randomized trees, and XGBoost are themselves ensembles, three additional ensembles are tested for comparison: a simple equal weighted average of the base learners, a stacked ensemble where the output of the base learners is fed into XGBoost, and a single-layer super learner.

\subsection{Benchmark Deep Neural Networks}
DNN are used to establish benchmarks for performance.  Some hyper-parameter tuning through experimentation is performed to achieve performance indicative of the capabilities of DNN. In Table~\ref{tab:2}, two DNN architectures are tested and described : a multi-layer perceptron (MLP) and a convolutional neural network (CNN) 

\begin{table}[!ht]
\centering
\caption{Summary of architecture and hyper-parameter values used for benchmark deep neural networks.}
\label{tab:2}
\begin{tabular}{@{}lll@{}}
\toprule
\multicolumn{1}{c}{Architecture} & \multicolumn{1}{c}{\begin{tabular}[c]{@{}c@{}}Hyper-parameters \\ (Multi-layer perceptron)\end{tabular}}                & \multicolumn{1}{c}{\begin{tabular}[c]{@{}c@{}}Hyper-parameters \\ (Convolutional neural network)\end{tabular}}              \\ \midrule
Convolutional layer              & N/A                                                                                                                     & \begin{tabular}[c]{@{}l@{}}Filters: 32; Kernel size: 5 or (5, 5); \\ Activation: RELU;\\ Weight constraint: 4\end{tabular} \\
Max pooling layer                & N/A                                                                                                                     & Pool size: 2 or (2, 2)                                                                                                      \\
Convolutional layer              & N/A                                                                                                                     & \begin{tabular}[c]{@{}l@{}}Filters: 16; Kernel size: 3 or (3, 3);\\ Activation: RELU; \\Weight constraint: 4\end{tabular}  \\
Max pooling layer                & N/A                                                                                                                     & Pool size: 2 or (2, 2)                                                                                                      \\
Dropout regularization           & N/A                                                                                                                     & Drop rate: 0.2                                                                                                              \\
Dense layer                      & \begin{tabular}[c]{@{}l@{}}Nodes: 128; Activation: RELU; \\ Weight constraint: 4\end{tabular}                         & \begin{tabular}[c]{@{}l@{}}Nodes: 128; Activation: RELU;\\ Weight constraint: 4\end{tabular}                              \\
Dense layer                      & \begin{tabular}[c]{@{}l@{}}Nodes: 64; Activation: RELU; \\  Weight constraint: 4\end{tabular}                         & \begin{tabular}[c]{@{}l@{}}Nodes: 64; Activation: RELU; \\ Weight constraint: 4\end{tabular}                              \\
Output layer                     & \begin{tabular}[c]{@{}l@{}}Nodes: number of classes; \\ Activation: Softmax\end{tabular}                               & \begin{tabular}[c]{@{}l@{}}Nodes: number of classes; \\ Activation: Softmax\end{tabular}                                   \\
Optimizer: Adam                  & \begin{tabular}[c]{@{}l@{}}Learning rate: 0.001; \\Learning rate decay: \\ Learning rate/$\sqrt{Max~epochs}$\end{tabular} & \begin{tabular}[c]{@{}l@{}}Learning rate: 0.001; \\Learning rate decay: \\ Learning rate/$\sqrt{Max~epochs}$\end{tabular}     \\
Batch size                       & 200                                                                                                                     & 200                                                                                                                         \\
Max epochs                       & 50                                                                                                                      & 50                                                                                                                          \\
Validation split                 & 0.2                                                                                                                     & 0.2                                                                                                                         \\
Early stopping patience          & 3                                                                                                                       & 3                                                                                                                           \\ \bottomrule
\end{tabular}
\vspace{-5mm}
\end{table}

\subsection{Datasets}
\subsection*{Sentiment Classification} The IMDB Movie reviews sentiment classification dataset contains 25,000 reviews for training and 25,000 for testing.  The reviews have been labelled as positive or negative~\cite{Maas:2011:LWV:2002472.2002491}.  The 2,000 most frequent words in the set are used to calculate the term frequency-inverse document frequency (TF-IDF) matrix.

\subsection*{Image Categorization}The MNIST database of handwritten digits is a commonly used dataset to test the performance of computer vision algorithms.  It includes a training set of 60,000 images and a test set of 10,000 images of handwritten digits 0 to 9.  The images are 28 pixels by 28 pixels in greyscale~\cite{Lecun1998Gradient-basedRecognition}.

\subsection{Performance Measures} 

Two metrics are used to evaluate the performance of the learning algorithms.  One is $Accuracy$, which is the proportion of correctly classified records, and the other is $Log Loss$. Both $Accuracy$ and $Log Loss$ formulas are shown in Equations~\ref{eq:1} and~\ref{eq:2}, respectively.
\begin{equation} \label{eq:1}
Accuracy = \frac{\sum_{x=1}^{n}\sum_{y=1}^{j}f(x,y)C(x,y)}{n}
=\frac{TP + TN}{TP + FP + TN + FN}
\end{equation}
Where $n$ denotes the number of instances, $j$ the number of classes, $f(x,y)$ the actual probability of instance $x$ to be of class $y$. $C(x,y)$ is one if and only if $y$ is the predicted class of $x$, otherwise $C(x,y)$ is zero. $Accuracy$ is equivalently defined in terms of the confusion matrix, where $TP$ is true positives, $TN$ is true negatives, $FP$ is false positives, and $FN$ is false negatives.
\begin{equation} \label{eq:2}
Log Loss = \frac{-\sum_{y=1}^{j}\sum_{x=1}^{n}f(x,y)log(p(x,y))}{n}
\end{equation} 
Where $f(x,y)$ is defined as above and $p(x,y)$ is the estimated probability of instance $x$ is class $y$.  Minimizing $Log Loss$, also known as cross entropy, is equivalent to maximizing the log likelihood of observing the data from the model.  Both $Accuracy$ and $LogLoss$ are commonly used performance measures in machine learning~\cite{Ferri2009}.

\section{Results}
\subsection{Sentiment Classification}
Log loss and accuracy results of DSL, base learners, benchmark ensembles, and benchmark DNN on the IMDB sentiment classification dataset are shown in Table~\ref{tab:3}.
\begin{table}[!ht]
\centering
\caption{Comparison of log loss and accuracy on IMDB test data}
\label{tab:3}
\begin{tabular}{@{}lrr@{}}
\toprule
Test                         & Log loss & Accuracy \\ \midrule
\textbf{Deep Super Learner (DSL)}  & \hspace{16mm}\textbf{0.28} & \hspace{16mm}$~~$\textbf{88.22\%}  \\
Multi-layer perceptron       & 0.29 & 87.53\%  \\
Super learner                & 0.30 & 86.59\%  \\
XGBoost stack ensemble       & 0.31 & 87.19\%  \\
Convolutional neural network & 0.31 & 86.40\%  \\
Logistic regression          & 0.32 & 87.78\%  \\
XGBoost                      & 0.39 & 84.45\%  \\
Simple average ensemble      & 0.42 & 86.57\%  \\
Random forest                & 0.46 & 84.27\%  \\
Extremely randomized trees   & 0.60 & 79.20\%  \\
k-Nearest neighbors          & 0.72 & 68.22\%  \\ \bottomrule
\end{tabular}
\end{table}

\begin{figure}[!ht]
\centering 
\includegraphics[width=12cm]{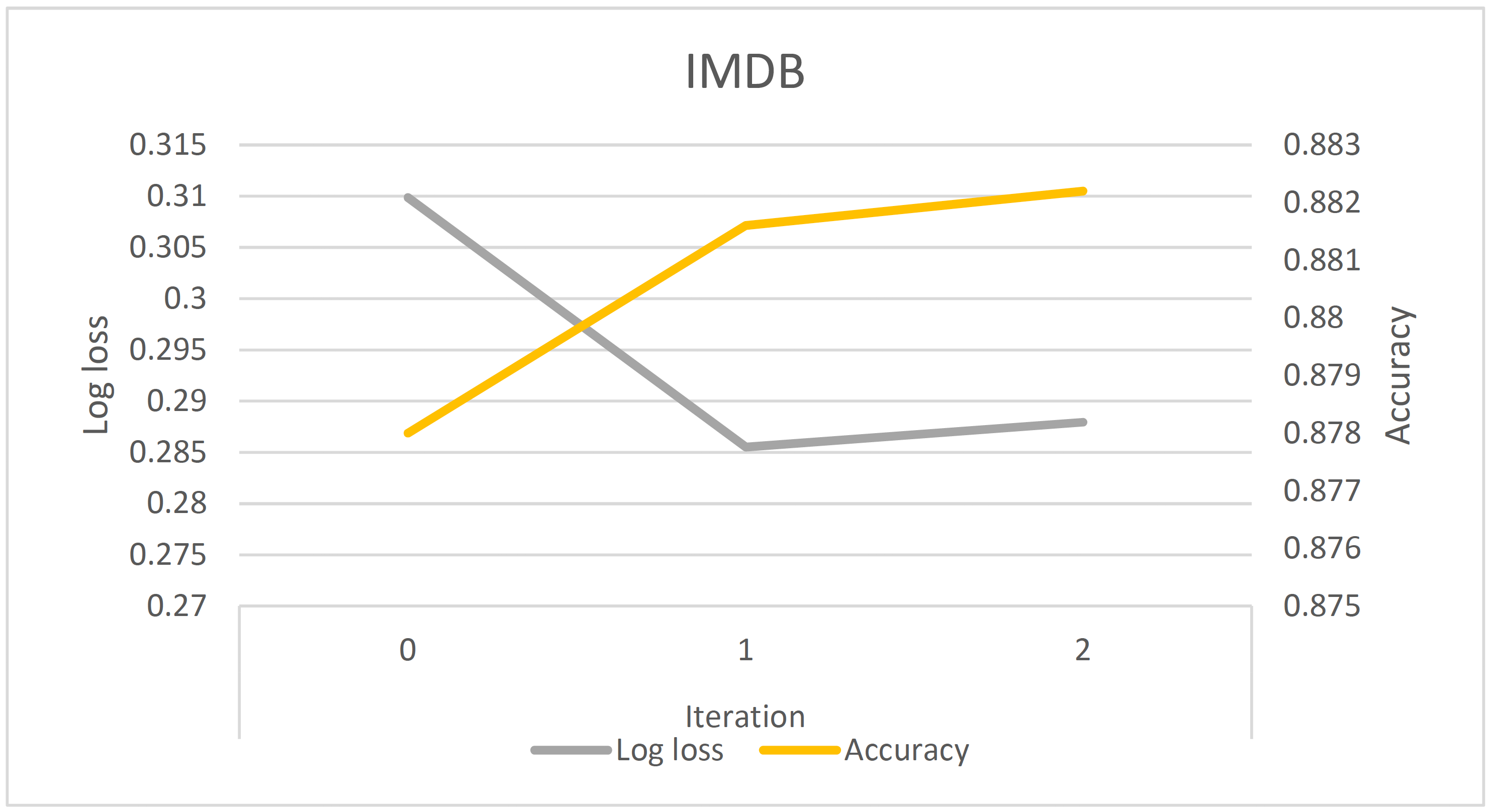}
\caption{Log loss and accuracy by iteration of the DSL on IMDB test data.} 
\label{fig:2} 
\end{figure}

The DSL achieved statistically significantly lower loss and higher accuracy than all other algorithms.  Since the TF-IDF matrix does not convey spatial or sequential relationships, DNN architectures CNN may not be expected to perform as well on this test.  The MLP, like DSL here, is set up to be more general purpose yet is outperformed by DSL.  DSL outperforming a single-layer super learner indicates adding depth to the algorithm improves performance.  Figure~\ref{fig:2} shows the performance of DSL on the IMDB test data by iteration.

\subsection{Image Categorization}
Log loss and accuracy results of DSL, base learners, benchmark ensembles, and benchmark DNN on the MNIST handwritten digits dataset are shown in Table~\ref{tab:4}.

\begin{table}[!ht]
\centering
\caption{Comparison of log loss and accuracy on MNIST test data.}
\label{tab:4}
\begin{tabular}{@{}lrr@{}}
\toprule
Test                         & Log loss & Accuracy \\ \midrule
Convolutional neural network & 0.03 & 99.17\%   \\
\textbf{Deep Super Learner (DSL)  }         & \hspace{16mm}\textbf{0.06} & \hspace{16mm}\textbf{98.42\% }  \\
Super learner                & 0.07 & 97.82\%   \\
Multi-layer perceptron       & 0.07 & 97.85\%   \\
XGBoost stack ensemble       & 0.08 & 98.24\%   \\
XGBoost                      & 0.08 & 97.65\%   \\
Simple average ensemble      & 0.18 & 97.65\%   \\
Random forest                & 0.24 & 97.00\%    \\
k-Nearest neighbors          & 0.26 & 96.68\%   \\
Logistic regression          & 0.27 & 92.55\%   \\
Extremely randomized trees   & 0.43 & 95.87\%   \\ \bottomrule
\end{tabular}

\end{table}

\begin{figure}[!ht]
\centering 
\includegraphics[width=12cm]{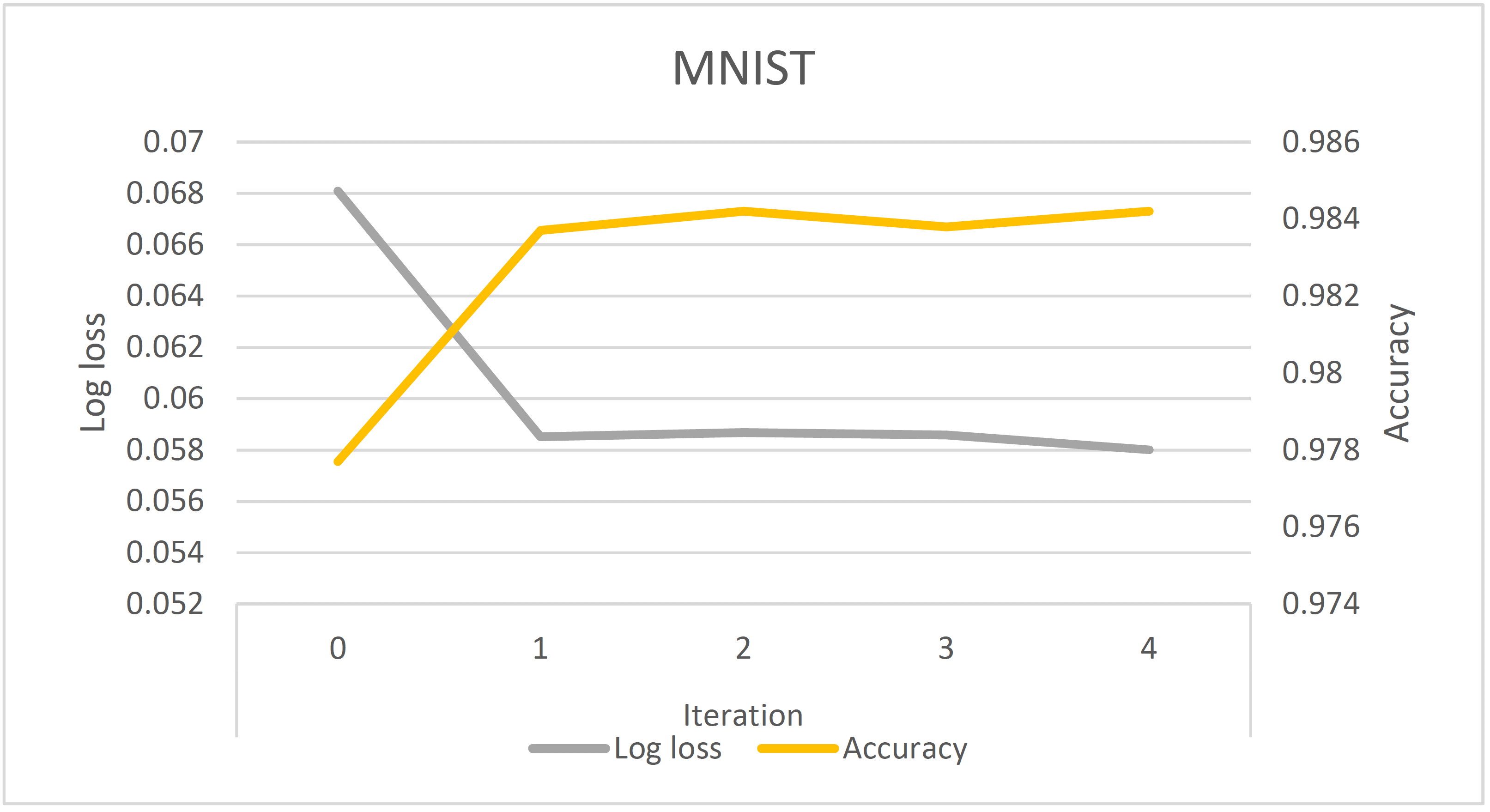}
\caption{Log loss and accuracy by iteration of the DSL on MNIST test data.} 
\label{fig:3} 
\end{figure}

The DSL achieved statistically significantly lower loss and higher accuracy than all algorithms except for CNN.  The design of CNN make them well suited to image processing.  Again, DSL outperformed MLP and super learner showing the advantages of diversity in learners and depth.  The order of the base learners by performance differs between the two datasets showing the importance of including a diverse set of learners when addressing various problems and the value of optimizing the component weights.    Figure~\ref{fig:3} shows the performance of DSL on the MNIST test data by iteration.

\subsection{Runtime}

All algorithms are implemented in Python using the scikit-learn library for logistic regression, k-nearest neighbors (KNN), random forest, and extremely randomized trees, XGBoost library for XGBoost, SciPy for the convex optimizer, and Keras with a TensorFlow backend for MLP and CNN.  Experiments are run on a desktop with an Intel Core i7-7700 with 16 GB of RAM.  DSL on IMDB converged after three iterations running for a total of 50 minutes, 46 of which are spent in the prediction phase of KNN.  MLP on IMDB converged after two epochs running for one minute.  CNN on IMDB converged after six epochs running for a total of seven minutes.  DSL on MNIST converged after five iterations, running for a total of 86 minutes, 70 of which are spent in the prediction phase of KNN.  MLP on MNIST converged in 12 epochs running for two minutes.  CNN on MNIST converged in 12 epochs running for a total of 12 minutes.  DSL is inherently parallel across component learners.  With optimized parallel processing and selection of base learners, the runtime of DSL can be dramatically reduced.

\subsection{Threats to Validity}

Threats to internal validity include errors in the experiments and implementations.  While the experiments and implementations were carefully double checked, errors are possible.  Threats to external validity include whether the results in this paper generalize to other datasets, sets of base learners and architectures.  While the tests include two rather different datasets, only one set of base learners and deep architecture were tested.  Applying the methods described here to additional datasets as well as varying base learners and architecture will reduce this threat.  Threats to construct validity include the appropriateness of the benchmark algorithms and evaluation metrics.  For the most part, benchmark ensembles outperformed their component learners and DNN outperformed ensembles of base learners on the same tasks as expected.  The use of log loss and accuracy are common in machine learning studies to evaluate the performance of prediction algorithms.

\section{Conclusion}

Results for the deep super learner are encouraging.  Using a weighted average of the base learners optimized to minimize log loss yields results superior to any individual base learner.  Using a cascade of multiple layers of base learners where each successive layer uses the output from the previous layer as augmented features for input to add depth to the learning improved performance further.  While still shy of the performance levels obtained by CNN on image data, the deep super learner using traditional machine learning algorithms outperformed MLP on image data and outperformed MLP and CNN on classification from a TF-IDF matrix while also having fewer hyper-parameters and providing interpretable and transparent results.  Though still in the early stages of development of a deep super learning ensemble, particularly compared to DNN, further development of the architecture, for example to better capture spatial or sequential relationships, should be conducted.

%
%
\bibliographystyle{splncs}      
\bibliography{canadianai}       
\end{document}